# Factors affecting the COVID-19 risk in the US counties: an innovative approach by combining unsupervised and supervised learning

**Samira Ziyadidegan[1]\*, Moein Razavi[2] \*, Homa Pesarakli[3]\*, Amir Hossein Javid[4], Madhav Erraguntla[5]**

| | |
|---|---|
| 1 | Department of Industrial and Systems Engineering, Texas A&M University, College Station, TX, 77843, samiraziyadg@tamu.edu. |
| 2 | Department of Computer Science and Engineering, Texas A&M University, College Station, TX, 77843, moeinrazavi@tamu.edu; |
| 3 | Department of Architecture, Texas A&M University, College Station, TX, 77843, Homa.pesarakli@tamu.edu; |
| 4 | Department of Statistics, Oklahoma State University, Stillwater, OK, 74074, Amir.javid@okstate.edu; |
| 5 | Department of Industrial and Systems Engineering, Texas A&M University, College Station, TX, 77843, merraguntla@tamu.edu. |
| 2 | Correspondence: moeinrazavi@tamu.edu; |
| \* | These authors contributed equally to this paper. |

**Abstract:** The COVID-19 disease spreads swiftly, and nearly three months after the first positive case was confirmed in China, Coronavirus started to spread all over the United States. Some states and counties reported high number of positive cases and deaths, while some reported lower COVID-19 related cases and death. In this paper, the factors that could affect the risk of COVID-19 infection and death were analyzed in county level. An innovative method by using K-means clustering and several classification models is utilized to determine the most critical factors. Results showed that *longitudinal coordinate* and *population density, latitudinal coordinate, percentage of non-white people, percentage of uninsured people, percent of people below poverty*, *percentage of Elderly people, number of ICU beds per 10,000 people, percentage of smokers* were the most significant attributes.

**Keywords:** multinomial logistic regression, K-means clustering, COVID-19, SARS-CoV-2, meteorological variables

## 1. Introduction

COVID-19 disease is caused by severe acute respiratory syndrome coronavirus 2 (SARS-CoV-2) with common symptoms of fever, dry cough, shortness of breath, and other signs of respiratory-related infections. World Health Organization (WHO) reported that 80% of patients experienced these symptoms mildly. However, older people (>60 years old) and persons with co-morbid diseases are at a higher risk for severe symptoms and death (Velavan and Meyer 2020; World Health Organization 2020). Besides, younger patients with no underlying disease might also experience severe symptoms or even death (Jahromi, Avazpour, et al. 2020; The Washington Post 2020; Yousefzadegan and Rezaei 2020).

The first positive case of COVID-19 in the United States was reported in the state of Washington on January 20, 2020. By March 17, 2020, Covid-19 has spread across all US states (Centers for Disease Control and Prevention 2020; Saad B. Omer, Malani, and Rio 2020). Figure 1 (plotted in Python using positive case numbers and death numbers in each county[1]; darker color corresponds to the higher number of cases (The New York Times 2021)) shows the aggregated COVID-19 positive case and death count maps for all US states until November 6, 2020. Reports showed that on November 6, 2020, the top states for positive COVID-19 cases are California, Texas, Florida, New York, and Illinois, while the top 5 states for death cases are New York, Texas, California, New Jersey, and Florida. (1point3acres 2020; John Hupkins University & Medicine 2020).

---

[1] primary legal divisions of most states in the U.S. (United States Sensus Bureau 2021)

Epidemiological models have been used for outbreak estimation and predicting upcoming peak and death rate. Accurate outbreak prediction can provide insight into problems caused by COVID-19 and be used to develop new policies (Pinter et al. 2020). For example, Erraguntla et al. (Erraguntla, Zapletal, and Lawley 2019) used a natural language algorithm and machine learning to develop a platform to analyze infectious disease regarding data integration, situational awareness visualization, prediction, and intervention. Some studies (Freeze, Madhav, and Verma 2018; Madhav et al. 2017) also developed Data Integration and Predictive Analysis System (IPAS) to integrate data in order to predict disease patterns, intensity, and timing using machine learning algorithms. The COVID-19 pandemic has shown a complex nature unlike other recent outbreaks (Ardabili et al. 2020). The COVID-19 can more rapidly spread through human contact comparing to other recent epidemics like Severe Acute Respiratory Syndrome (SARS) and Middle East Respiratory Syndrome (MERS) (Mallapaty 2020). Furthermore, many known and unknown variables affect in the spread of COVID-19. So, it has an uncertain out-break prediction that cannot be estimated accurately by standard epidemiological models. The growing big data of the number of infected subjects, death counts, and possible influential factors requires to be managed and analyzed by innovative solutions (Sujath, Chatterjee, and Hassanien 2020). Multiple sources of data provide information about the various region across the county including the growth of infection, geographical, local services and policies, and demographical data. In order to have more precise and long-term prediction, Ardabili et al. (Ardabili et al. 2020) showed that the application of various Machine learning algorithm is more effective in estimating covid-19 positive case rate and death rate. They also mentioned that estimating the factors that affect the death rate is important in estimating the number of patients and planning new facilities.

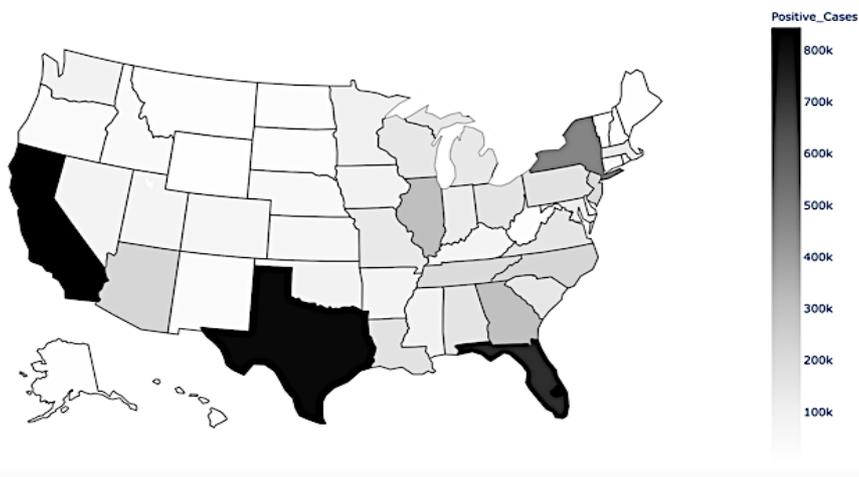

(a)

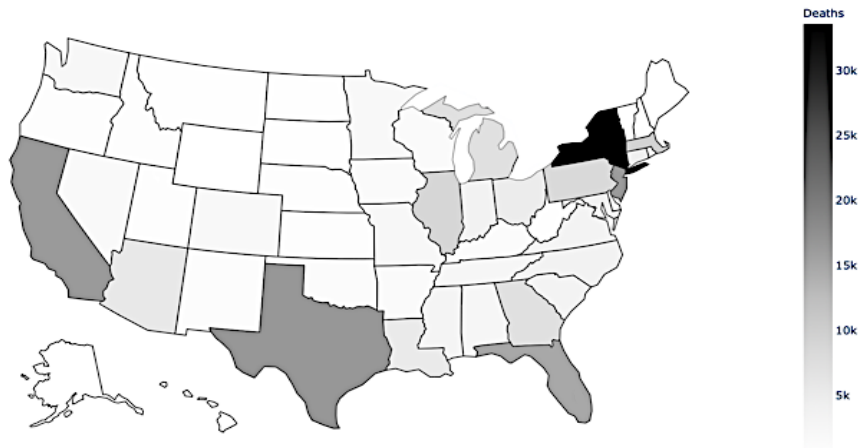

(b)

*Figure 1 COVID-19 positive case and death counts maps for all states of the US, (a) COVID-19 positive case counts map, (b) COVID-19 Death counts map*

Previous studies have been shown that several factors affect COVID-19 spread and death rate. They showed that meteorological factors could have an essential role in the spread of a virus (Chien and Chen 2020; Wang et al. 2010). Before, emerging of COVID-19 the effect of meteorological factors also was tested on spread of other infectious diseases (Madhav et al. 2010). Climatic factors like humidity, sunlight, temperature, and wind speed can affect the droplet stability in the environment and thus affect COVID-19 transmission (B. Chen et al. 2020). Temperature can affect the life cycle and proliferation of viruses; precipitation and rainfall can impact water-borne virus diffusion, while humidity and the wind speed can affect the dissemination of air-borne viruses (P. S. Chen et al. 2010; Lowen et al. 2007; Pica and Bouvier 2012; Wang et al. 2010; Wu et al. 2016). Different studies showed contradictory effects regarding the meteorological attributes on the transmission or lifespan of Coronavirus. Liu et al. (Liu et al. 2020) studied the correlation between confirmed case counts and weather-related attributes. The results showed that diurnal temperature, ambient temperature, and humidity are significant attributes and have negative correlations with the transmission of COVID-19. Moges Menebo (Menebo 2020) found that the maximum temperature and the normal temperature (average of high and low temperatures) have positive, but precipitation has negative impact on COVID-19 case counts. Ma et al. (Ma et al. 2020) evaluated the effects of meteorological and air pollutant factors on COVID-19 death rates. They concluded that diurnal temperature range and absolute humidity are positively and negatively associated with COVID-19 death rate, respectively. He et al. (He et al. 2021) studied the effects of temperature and relative humidity on daily COVID-19 confirmed positive cases in different Asian cities. Their study revealed that in different cities, the impact of temperature on COVID-19 confirmed cases could differ. In Beijing, Shanghai, and Guangzhou, the correlation between temperature and the number of positive cases was negative, while in Japan, it was positive. The study conducted by Pahuja et al. (Pahuja et al. 2021) demonstrated a positive correlation between the wind speed and the COVID-19 case counts. In contrast, Coccia (Coccia 2020) showed that the high wind speed could reduce the number of COVID-19 cases. Pramanik et al. (Pramanik et al. 2020) evaluated the effect of climatic factors on increasing COVID-19 cases in the climatic region of Russia. This study found that temperature seasonality has the highest association with the spread of COVID-19 in the humid continental region while, in the sub-arctic region, both daily and temperature seasonality had the highest effect on COVID-19 transmission. The conclusions regarding the relationship between climatic factors and COVID-19 transmission are still not conclusive.

In addition to the meteorological factors, there are other factors that impact the transmission and spread of COVID-19 positive case and death case counts. They include air pollutant factors such as the concentration of air pollutants (e.g., $PM_{10}$ and $SO_2$) (Coccia 2020; Ma et al. 2020), demographic factors (e.g., elderly people index, population density, etc.) (Coccia 2020; Figueroa et al. 2021), adults smoking history (Patanavanich and Glantz 2020; Zheng et al. 2020), and disease-related factors (e.g., cardiovascular or respiratory disease experience) (Bansal 2020; Coccia 2020; Jordan, Adab, and Cheng 2020; Zheng et al. 2020). Besides, the COVID-19 outbreak challenges medical systems worldwide in many aspects like increasing the demands for hospital beds and medical equipment (Zoabi, Deri-Rozov, and Shomron 2021). The shortage of the healthcare resources creates challenges for treating critically ill patients (Shoukat et al. 2020).

Reviewing the literature indicated that an analysis that took into account a comprehensive set of factors affecting the rate of positive COVID-19 cases and its death rate was missing. The contradictory findings might be due to confounding factors and a comprehensive study based on a wide range of predictor variables will result in more accurate characterization. Besides, as Figure 1 shows, states with a high number of COVID-19 cases (e.g., Florida, New York, and California) are located in areas with different geographical, health, demographic, and meteorological characteristics. However, the majority of the previous studies analyzed the factors affecting COVID-19 transmission in the areas with similar characteristics (e.g., geographical, demographical, etc.). Several studies evaluated the parameters that influence the COVID-19 case rate and death rate on the county level across the US. For example, Figueroa et al. (Figueroa et al. 2021) assessed the effect of race, ethnicity, and community levels of COVID-19 case rates and death rates in US counties. Li et al. (Adam Y. Li et al. 2020) also evaluated the effect of various risk factors across all US counties on a specific day. To the authors' knowledge there is no comprehensive model that considers more extensive areas (e.g., all US counties) with different geographical, health, demographic, and meteorological features and in a broader time frame is missing. This model will be able to characterize the differences in risk across these diverse regions.

In this study, a comprehensive set of factors that affect the risk level of COVID-19 in all US counties have been analyzed In the United States, a county is a subdivision of a state with specific geographical boundary that is used for political and administrative purposes and has some governmental authority (National Association of Counties 2021). To perform this analysis, we combined unsupervised and supervised learning using clustering analysis and classification models. The results of this study show why some areas have higher risk of COVID-19 than other areas.

This paper is organized as follows. In Section 2, the methodology is described. Section 3 introduces the datasets used in this paper. Then, the process of data preparation and cleaning is explained. The analysis results are presented in Section 4. Finally, the findings are discussed in Section 5.

## 2. Methodology

*Methodology*

Figure 2 indicates the flow used in this article to determine the significant factors affecting COVID-19 transmission and the risk level of each county. To perform the data preparation step, since different datasets were given from the different sources (which are addressed in the Data Availability Statement section), they should be combined, and the inaccurate and unmatched records should be corrected. Next, for feature reduction and find the highly correlated parameters, we calculated the Pearson correlation coefficients between different parameters, and chose a single variable from each set of highly correlated variables (where the absolute correlation value is higher than 0.7).

In the next step, clustering analysis is applied on COVID-19 related data (positive case rates and death rates) K-means clustering method. The Elbow method is used to find the proper number of clusters. Using the clustering results, we labeled the counties into different classes of risk levels. In order to profile the clusters and understand the features affecting cluster membership (and therefore, the features determine the risk level of counties), we applied different classification models on the cluster labels.

Moreover, to overcome the class imbalance (different clusters contain different number of counties), we applied Synthetic Minority Oversampling Technique (SMOTE) (Chawla et al. 2020). The classification model with the best performance accuracy was selected to determine the significant variables affecting COVID-19 risk levels using Mean Decrease Accuracy, Mean Decrease in Gini and SHapley Additive exPlanation (SHAP) values. After determining the significant variables, we evaluated their effects on the counties' risk-level.

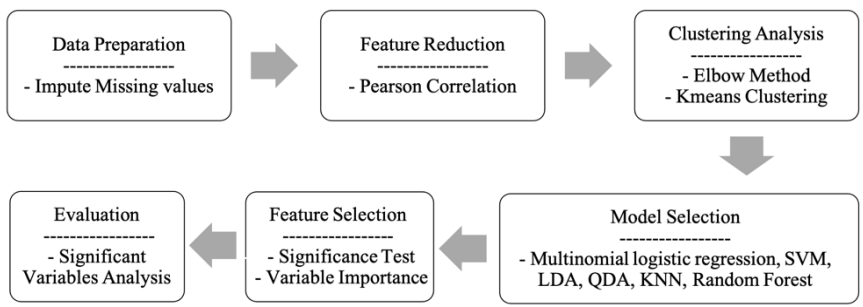

*Figure 2 Methodology used in the study*

## 3. Data Preparation

### 3.1. Data Collection

The data used in this study was taken from various online sources from March 2020 to December 2020, which includes COVID-19 positive cases and deaths, demographic, meteorological, health, and location-based data for each US county (Table 1). The source for each dataset is given in the Data Availability Statement section at the end of the paper. The COVID data is extracted until December 11th, 2020, when the first COVID vaccine was shot (Guarino et al. 2020). To evaluate the risk level of each county, we used the COVID-19 positive cases and death rates (positive cases and death counts divided by the population of the county) instead of their actual numbers. In this study, we used Building America (BA) climatic regions by county as meteorological data. Eight climate regions are identified, including hot-humid, mixed humid, hot-dry, mixed dry, cold, very-cold, subarctic, and marine (Baecheler et al. 2010). Percentage of people with obesity (BMI > 30 (Ogden et al., 2020)), percentage of people below poverty (people with income lower than the threshold determined by the United States Census Bureau (United States Census Bureau 2020)), population density along with the percentage of elderly people (>65 years old) and percentage of non-white people were also included in the dataset. The parameters used in this study are shown in Table 1.

Table 1 List of all the parameters used in the study

| Category | Parameters |
|---|---|
| COVID-19 | positive rate, death rate |
| location-based | Longitudinal coordinates, latitudinal coordinates, percent of rural areas |
| meteorological | BA Climate Zone |
| Health | Number of ICU beds per 10,000, percent of smokers, percent of adults with obesity, percent of people uninsured, percent of adults with diabetes |
| Demographic | Percent of elderly population, percent of non-white population, percent of people below poverty, population density |

*3.2. Data Cleaning*

After combining the data from various sources, it was found that some values were missing in different variables. Missing values for the demographic data were replaced with data from the United States Census Bureau website (United States Census Bureau 2020). If no data was available for a specific county, values were imputed using the average of the non-missing values of that parameter across the state for income-related data. For Climate zone data, the missing values are imputed with the data available from one neighboring county. The total number of counties analyzed in the current study was 3127.

*3.3. Checking Correlations*

After normalizing all independent parameters, Pearson Correlation was calculated for all variables (correlation matrix shown in Figure 3). All parameters are kept since none of the correlation values are above 0.7 (the threshold for significant collinearity).

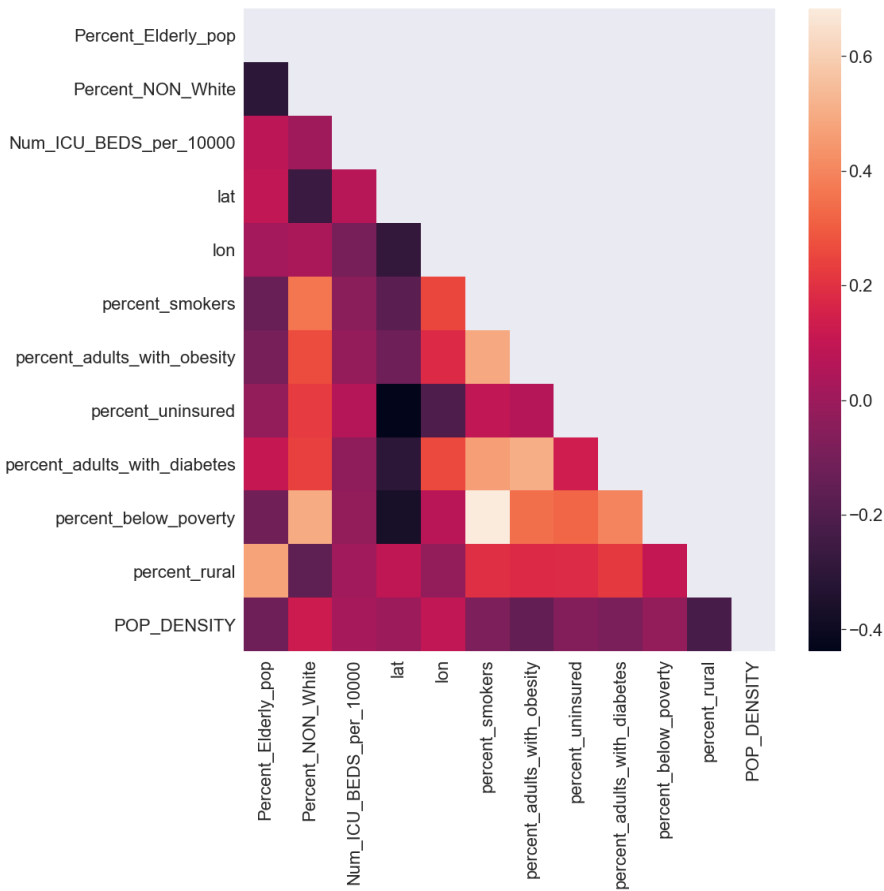

Figure 3 Correlation matrix for all parameters

## 4. Results

*4.1. Clustering Analysis*

      K-means clustering was performed on COVID-19 positive rates and death rates to determine the COVID-19 risk level for each county. Clustering was performed to group the counties based on similarities in their risk profile. The Elbow method (Figure 5) is used to determine the optimal number of clusters that can define the risk level of each county. The Elbow method is a visual method that can determine the optimal number of clusters considering the total within-cluster sum of squares of Euclidean distances (the cost). The optimized k value (k is the number of clusters) is such that adding another cluster (k+1) does not significantly decrease the benefit. In the Elbow method plot, the optimal k value is located at the elbow of the curve (Kodinariya and Makwana 2013; Purnima and Arvind 2014).

Figure 4 shows that k=3 could be the optimal number of clusters. The selected number of clusters (k =3) has a small SSE. As *k* increases, the SSE reduces toward zero, and the SSE value is zero when all data points have their own cluster, and k gets to infinity. However, the goal is to select the lowest SSE as *k* value is still small.

      Figure 5 shows the K-means clustering result that clustered the counties into three groups of low positive-low death rate, medium positive-medium death rate, and high positive-high death rate. The county located far apart from the other points (in the top right quadrant in Figure 5) belongs to the New York County, with a positive rate of 21.51% and a death rate of 1.5%.

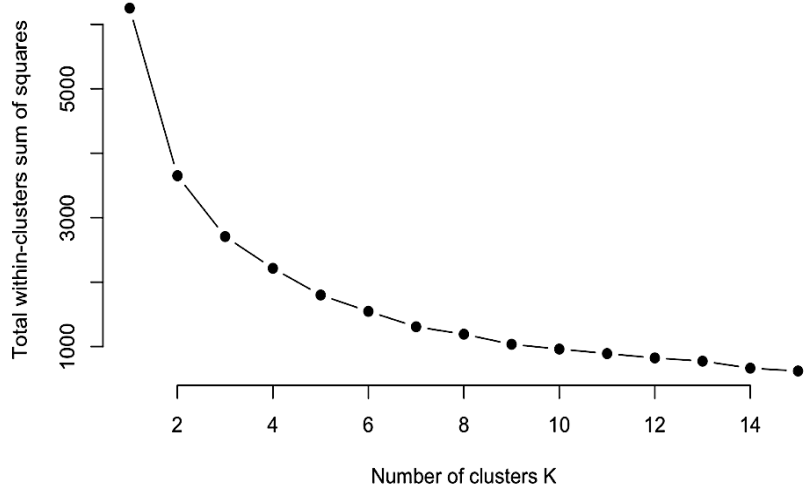

Figure 4 Elbow Method

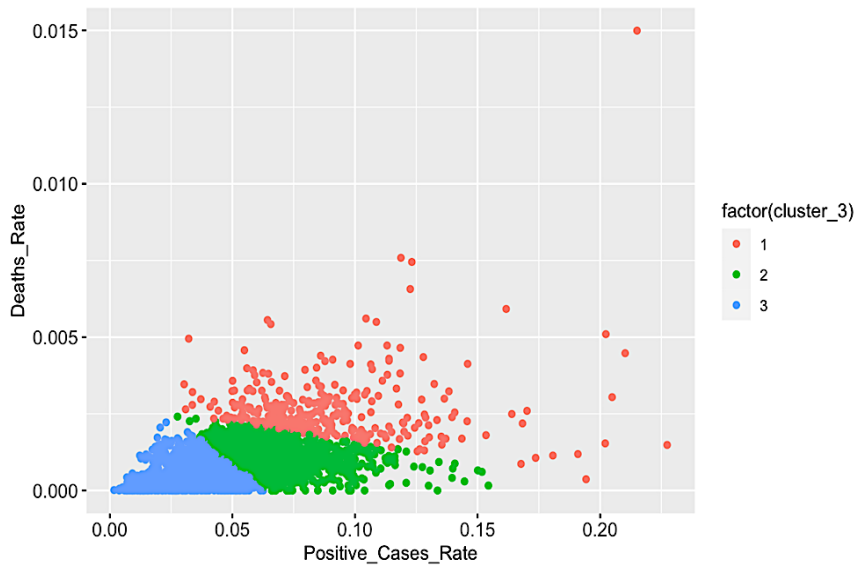

Figure 5 Clustering output plot

Figure 6 and Table 2 show descriptive results of each cluster. Cluster 3 is the low-risk cluster. It contains the highest number of counties with the lowest positive rate and death rate on average. On the other hand, cluster 1 has the lowest number of counties but the highest positive rate and death rate on average, which refers to high-risk counties. Cluster 2 is considered as the medium-risk cluster.

In the next section, classification analysis is applied to the results of clustering to find the significant parameters that affect the risk level of each county. Three COVID-19 risk levels of Low, medium, and high were used as labels for the classification analysis.

Table 2 Cluster attributes

| Cluster number | Counts | Mean Positive Rate | SD Positive Rate | Mean Death Rate | SD Death Rate |
|---|---|---|---|---|---|
| 1 | 306 | $8.94 \times 10^{-2}$ | $2.43 \times 10^{-2}$ | $2.68 \times 10^{-3}$ | $0.07 \times 10^{-2}$ |
| 2 | 1293 | $6.78 \times 10^{-2}$ | $1.60 \times 10^{-2}$ | $1.04 \times 10^{-3}$ | $0.05 \times 10^{-2}$ |
| 3 | 1528 | $3.50 \times 10^{-2}$ | $1.91 \times 10^{-2}$ | $5.14 \times 10^{-4}$ | $0.05 \times 10^{-2}$ |

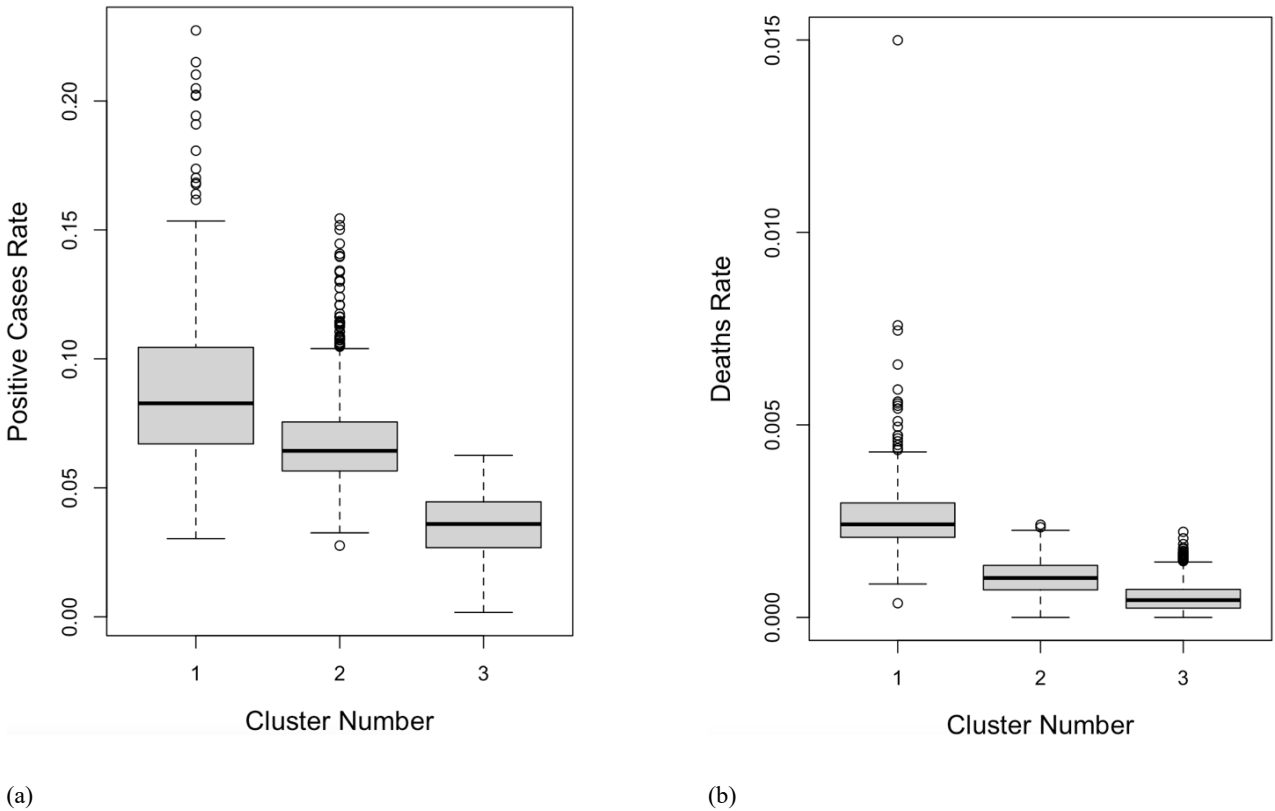

(a)  (b)

Figure 6 Clustering Results: (a) Positive Case Rate; (b) Death Rate

*4.2. Model Selection*

As can be seen in Table 2, there is size imbalance between different classes. We addressed the class imbalance using SMOTE. Then, to determine the significant factors, different classification models were employed to characterize COVID-19 risk clusters. Based on the accuracy values attained, the best model was used to select the factors with the highest significance in the transmission and death rate of COVID-19.

The data was divided randomly into train and test sets (80% and 20%, respectively) for classification. We applied 10-fold cross validation on the training dataset. Table 3 shows the classification models used in this study and their respective cross-validation and test accuracies. Among the classification models, Random Forest obtained the best performance on the test data. The linear models including MLR, LDA, and SVM Linear performed similarly. Therefore, to perform feature selection, we select the Random Forest model.

Table 3 Cross validation and test accuracy of the classification models used

| Method | Cross Validation Accuracy | Test Accuracy |
|---|---|---|
| Multinomial Logistic Regression | 59.93% | 60.99% |
| LDA | 59.55% | 60.90% |
| QDA | 49.96% | 48.07% |
| KNN | 83.16% | 76.86% |
| SVM Linear | 60.36% | 62.24% |
| SVM Radial | 94.13% | 80.18% |
| SVM Polynomial | 94.10% | 79.10% |
| **Random Forest** | **100%** | **82.15%** |

*4.3. Feature Selection*

The Random Forest model was used to identify the parameters which affect a county's COVID-19 risk level. Figure 7 shows the variable importance scores of Mean Decrease Accuracy (MDA, a measure of model accuracy loss by excluding each variable) and Mean Decrease in Gini (MDG, a measure of each variable's contribution to the homogeneity of the nodes and leaves in the resulting Random Forest) (Han et al., 2016). Figure 8 and 9 show SHAP value (Lundberg and Lee 2017).

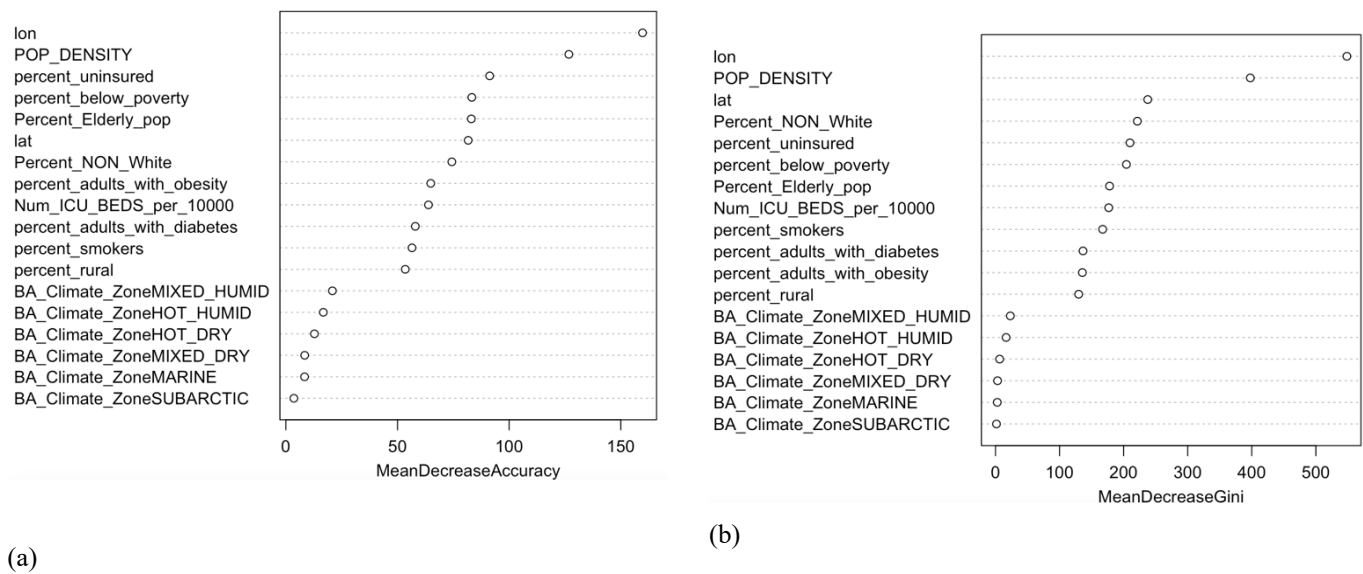

(a)         (b)

Figure 7 Feature importance plots for Random Forest model by (a) The Mean Decrease Accuracy (b) Mean Decrease in Gini

Based on both MDA and MDG criteria, *longitudinal coordinate* and *population density* are the highest contributing factors to the level of COVID-19 transmission and death rate. In the next level, *latitudinal coordinate, percentage of non-white people, percentage of uninsured people, percent of people below poverty*, *percentage of Elderly people, number of ICU beds per 10,000 people, percentage of smokers* are the significant variables. The significance of these parameters is confirmed using the SHAP values (Figures 8 and A1). Figure A1 reveal how the likelihood of the belonging of each variable to each cluster (risk level) varies by different values of that variable. We have brought an interpretation for the important variables found by SHAP values, Mean Decrease Accuracy, Mean Decrease in Gini in the discussion section.

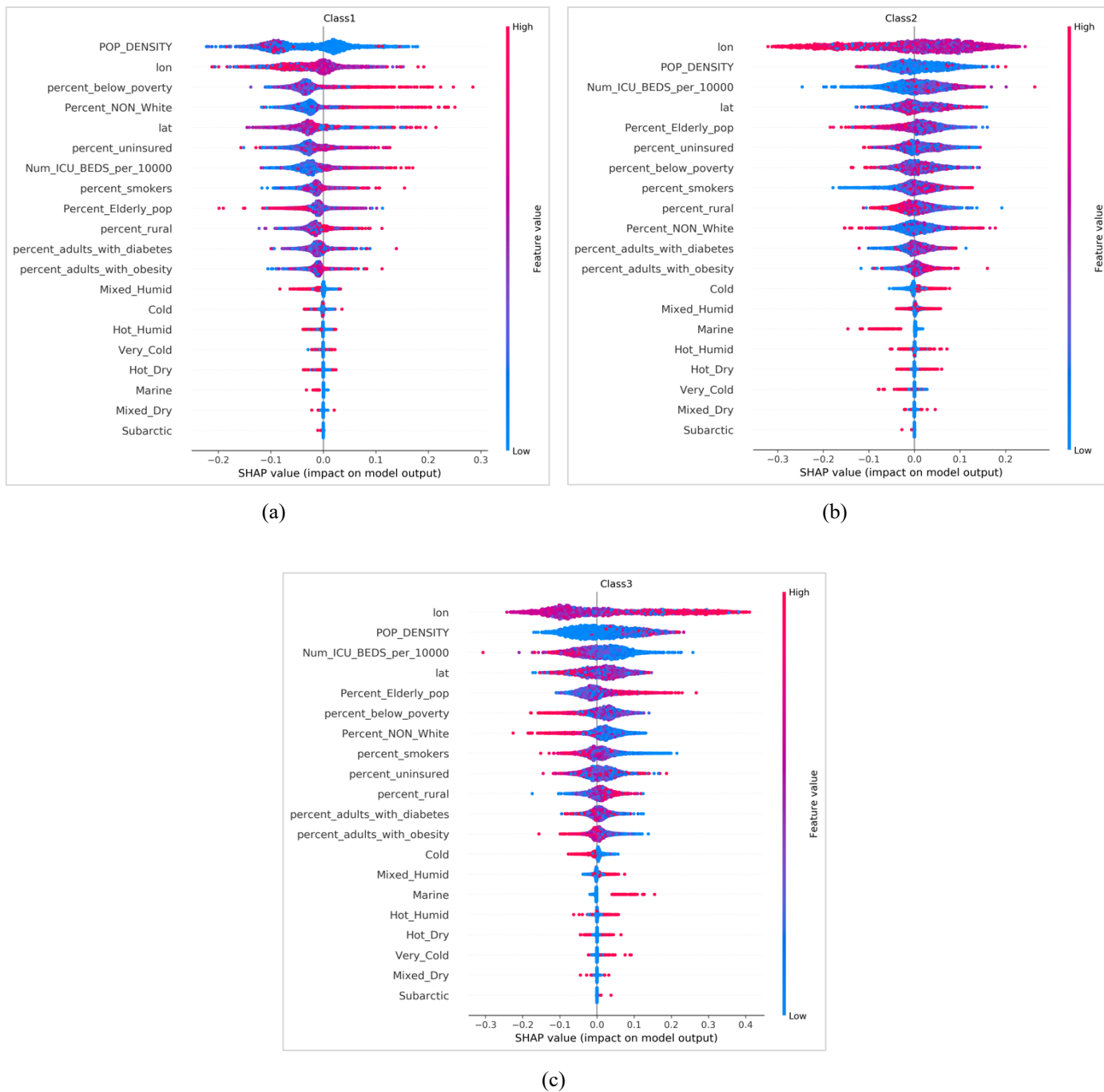

Figure 8 SHAP-values for (a) Class 1 (High-Risk Cluster), (b) Class 2 (Medium-Risk Cluster), (c) Class 3 (Low-Risk Cluster)

## 5. Discussion

In this paper, we analyzed demographic, meteorological, geographical, and health factors in determining the critical parameters affecting the transmission and death level of COVID-19 in the US counties. Aggregated COVID-19 positive cases and death counts for each county were derived, and their rate with respect to population was calculated. Pearson Correlation was used to test for collinearities and no variable was removed due to collinearity. Next, K-means clustering was applied to group the counties based on COVID-19 positive rates and death rates. According to the Elbow method, three clusters were chosen, representing low, medium, and high-risk clusters. The levels obtained from clustering were then considered as the nominal dependent variable for classification.

Several classification models were applied to the data, using cluster labels. Random Forest was finally chosen for a more accurate risk-level prediction and better interpretation of the effect of factors.

Parameters of *longitudinal coordinate* and *population density, latitudinal coordinate, percentage of non-white people, percentage of uninsured people, percent of people below poverty*, *percentage of Elderly people, number of ICU beds per 10,000 people, percentage of smokers* are the most significant ones using Mean Decrease Accuracy, Mean Decrease in Gini, and SHAP values (Figures 7, 8 and A1).

Considering longitudinal coordinates, counties located within central states (e.g., Harris County in Texas) have a higher chance to belong to the higher risk class. On the other hand, counties located within central latitudinal coordinates have a lower chance of belonging to the higher risk class (Studies showed that COVID-19 transmission is dependent upon seasonal dynamics. Longitude is a factor that correlates with seasonal dynamics and affects the COVID-19 transmission (Keshavarzi 2020; Skórka et al. 2020)).

Results reveal that counties with higher percentage of poverty, percentage of people uninsured, percentage of non-white people, percentage of smokers and the number of ICU beds per 10,000 people have more are more associated with a cluster with a higher level of COVID-19 risk.

Low-income people might have limited access to health products such as masks and sanitizers, which affects virus transmission and death (Jahromi, Mogharab, et al. 2020; Ramesh, Siddaiah, and Joseph 2020). They are less likely to work from their homes due to unstable jobs and income or less likely to have reliable and valid information about the COVID-19 (Little et al. 2021; Patel et al. 2020).

Additionally, low-income people are less likely to have health insurance (the higher percentage of people uninsured). So, due to high medical expenses, they prefer not to go to clinics/hospitals or use medications, which might increase the COVID-19 death rate and, as a result, increase the association of a county to a higher risk cluster.

The higher percentage of non-white people in a county increases the chance that the county belongs to a higher risk cluster. The reasons could be the lower educational level (so, these people are more likely to work non-remotely and being exposed to the COVID-19 cases (MLA Daly, Shelby R. Buckman, and Lily M. Seitelman 2020)) and lower number of USA-citizens who are less insured (Figueroa et al. 2021; Sommers et al. 2020).

A higher percentage of smokers in a county would increase the chance of belonging that county to the higher COVID-19 risk. This finding is in line with the previous studies (Patanavanich and Glantz 2020; Puebla Neira et al. 2021; Zheng et al. 2020); smokers are more likely to be hospitalized or die from COVID-19.

Results show that in counties with higher risk of COVID-19, the number of ICU beds are higher. We predict that the higher number of cases is mostly related to the counties with higher quality of health care and medical equipment.

Results indicates that counties with dense population have a higher chance of being in higher risk clusters. This finding is in line with similar studies (Rocklöv and Sjödin 2021). In dense areas, people cannot keep physical distance from others which is one of the most important factors to prevent the transmission of COVID-19 (Centers for Disease Control and Prevention 2021; Moein Razavi et al. 2021).

Results show that higher percentages of elderly people are associated with counties with lower COVID-19 positive case rates and death rates. Although older people are more likely to be hospitalized or die from this disease (Velavan and Meyer 2020; World Health Organization 2020), we predict that they are more likely to be quarantined and keep physical distances. So, there is less chance for them to be exposed by COVID-19.

**Appendix**

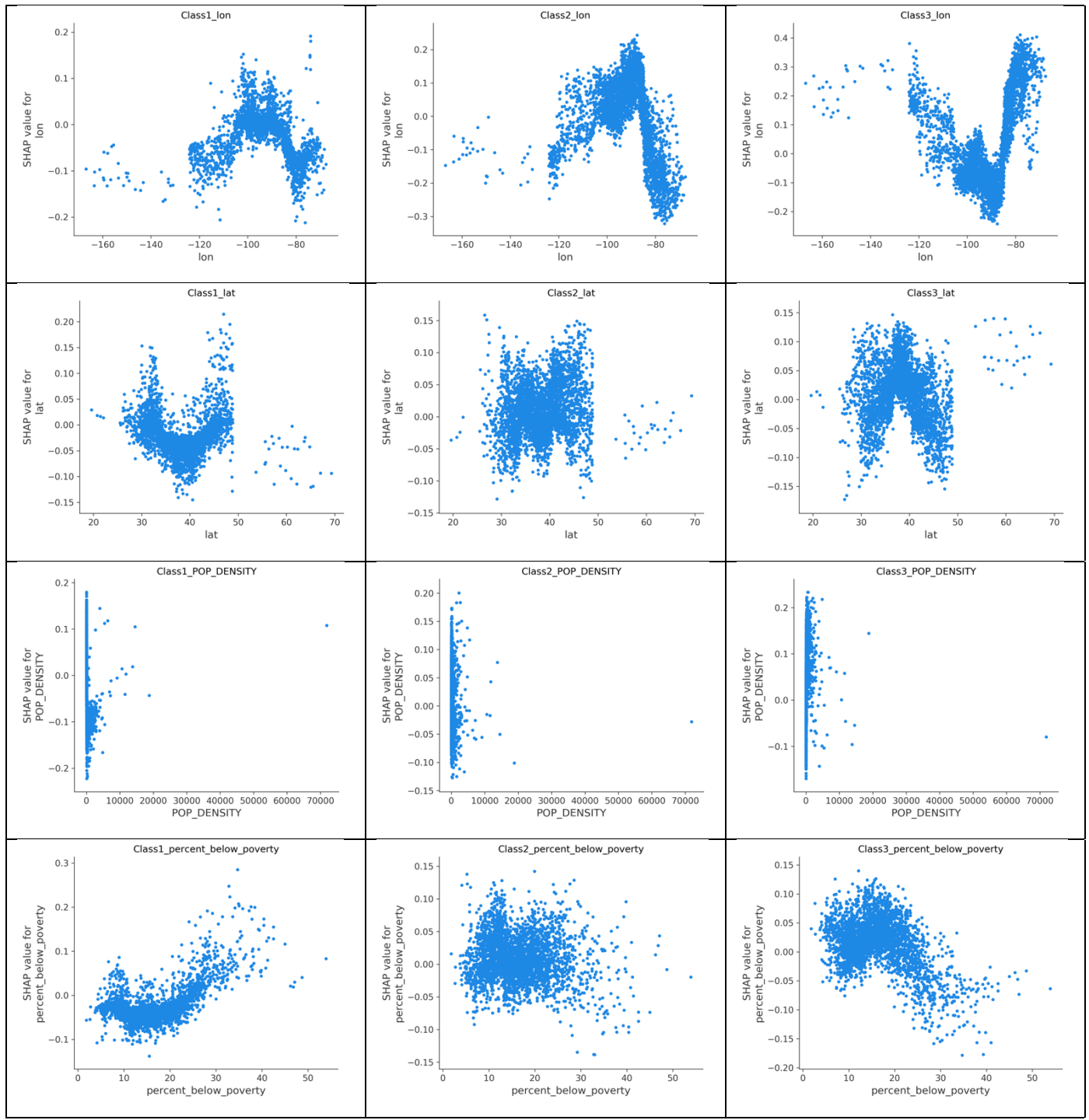

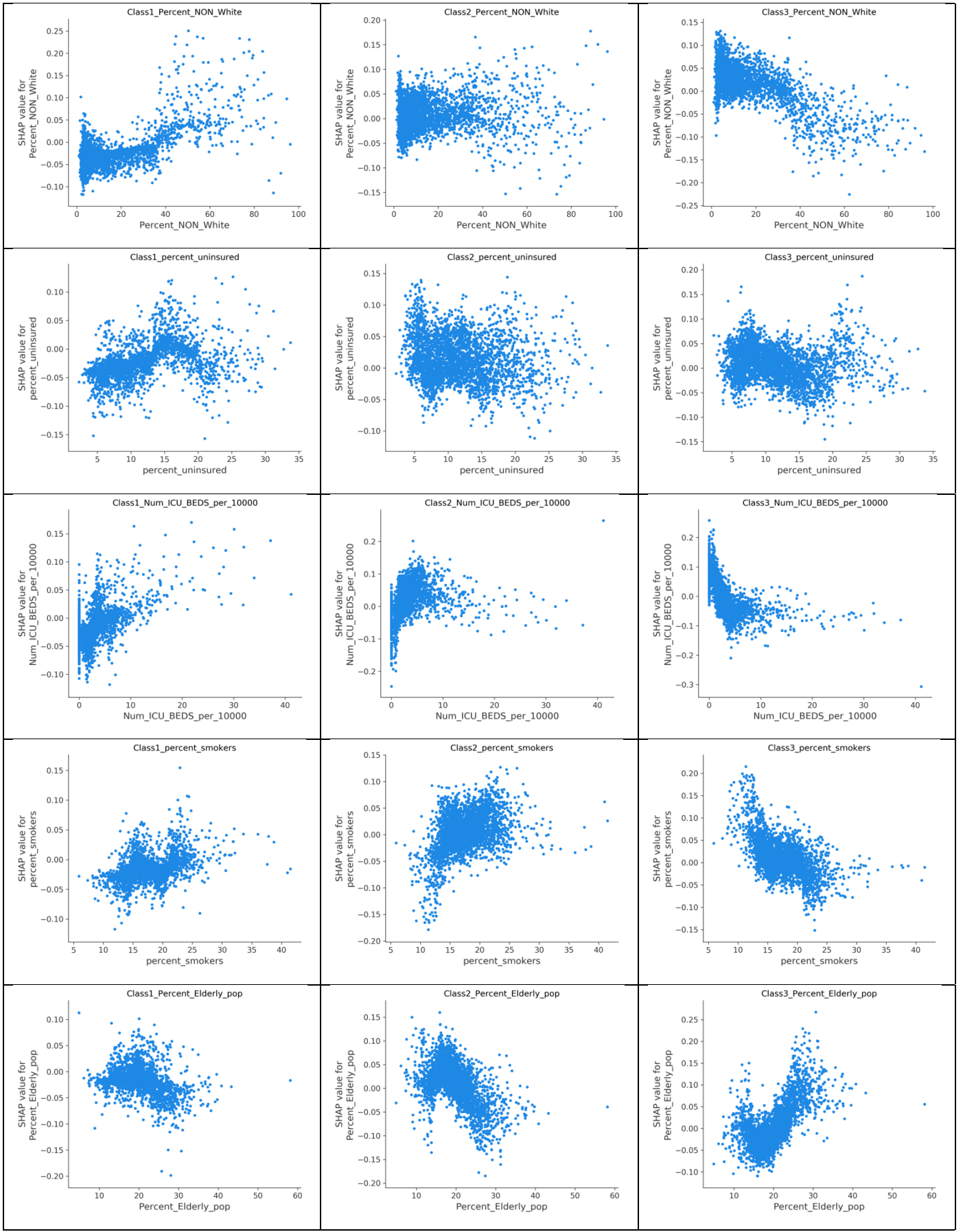

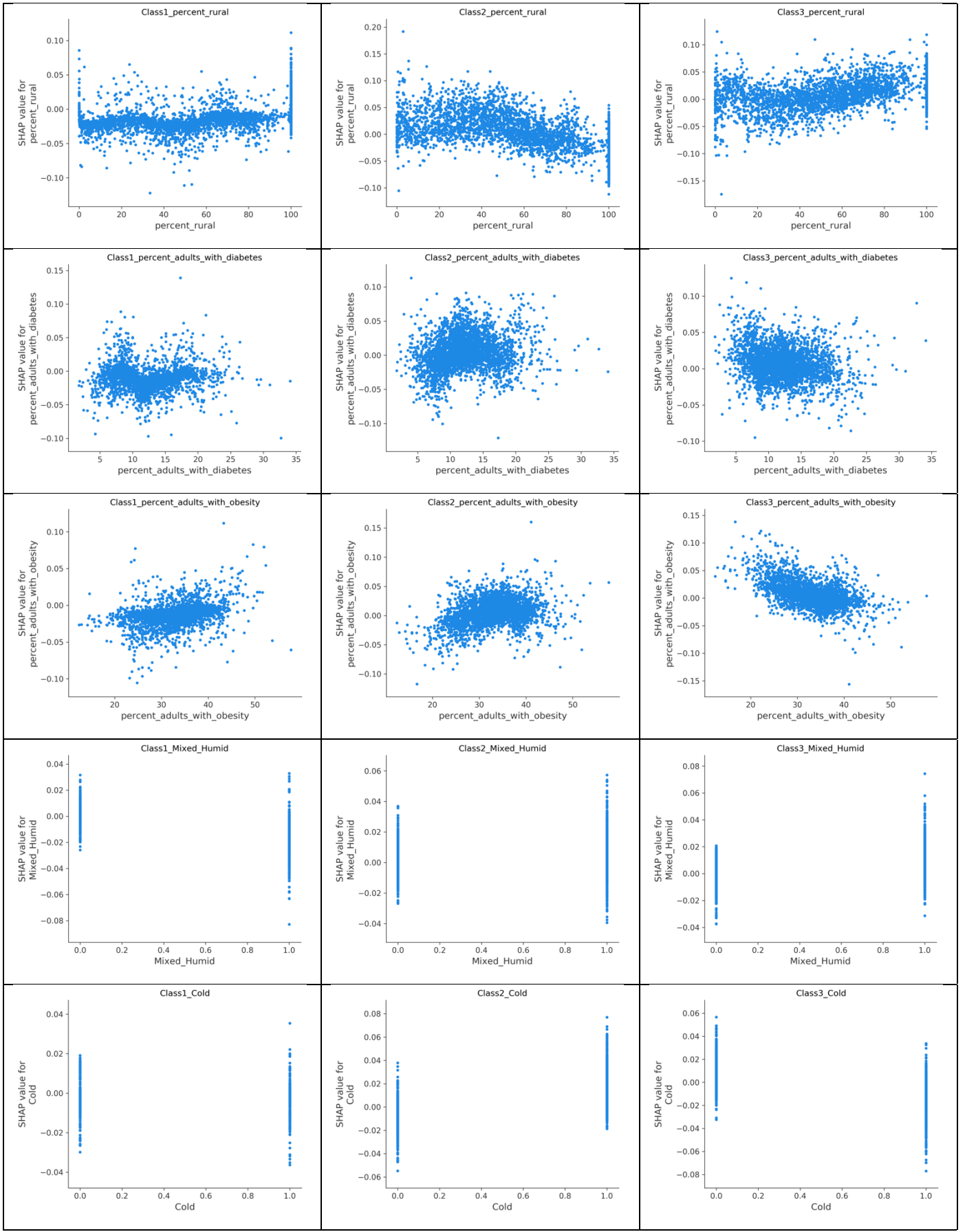

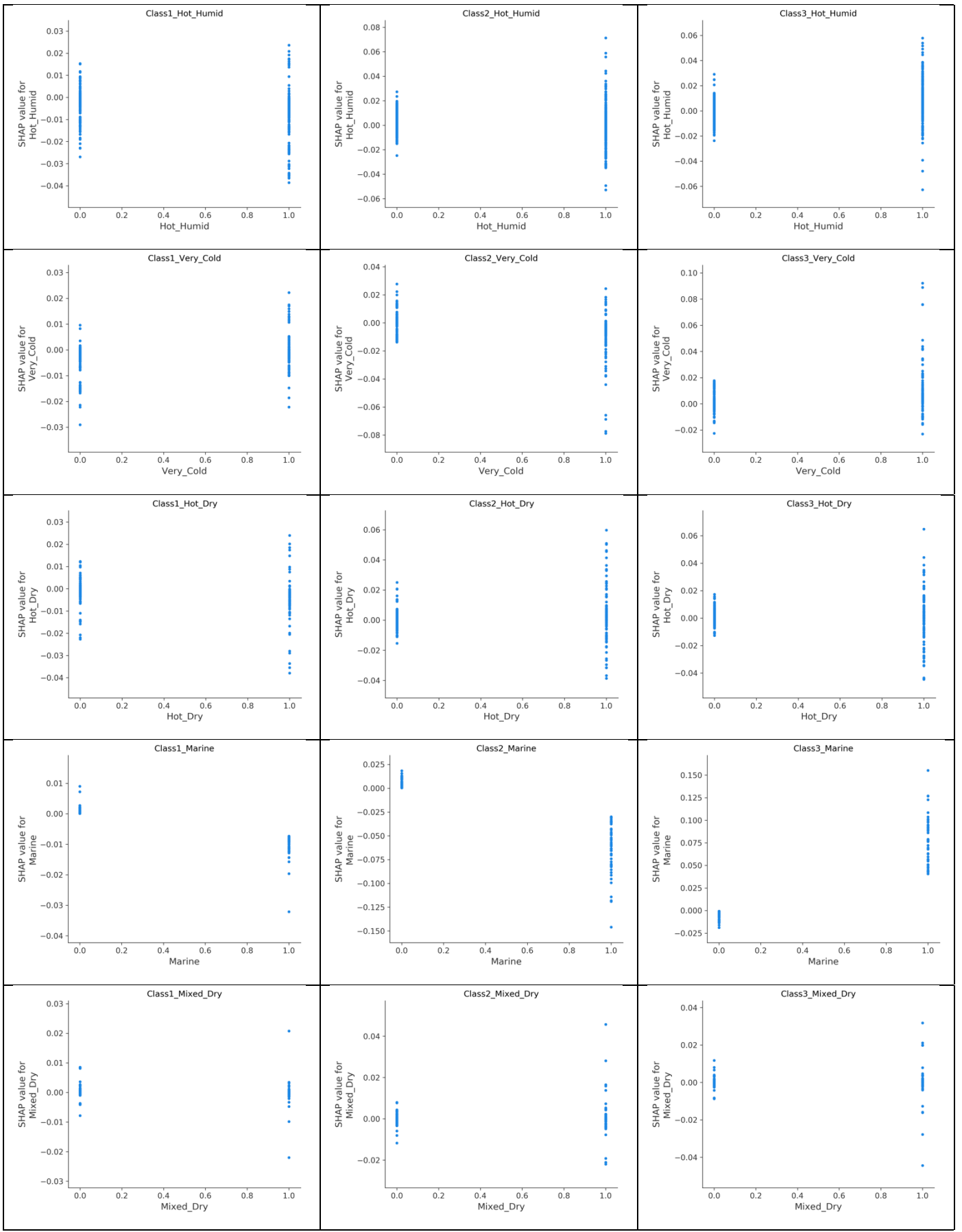

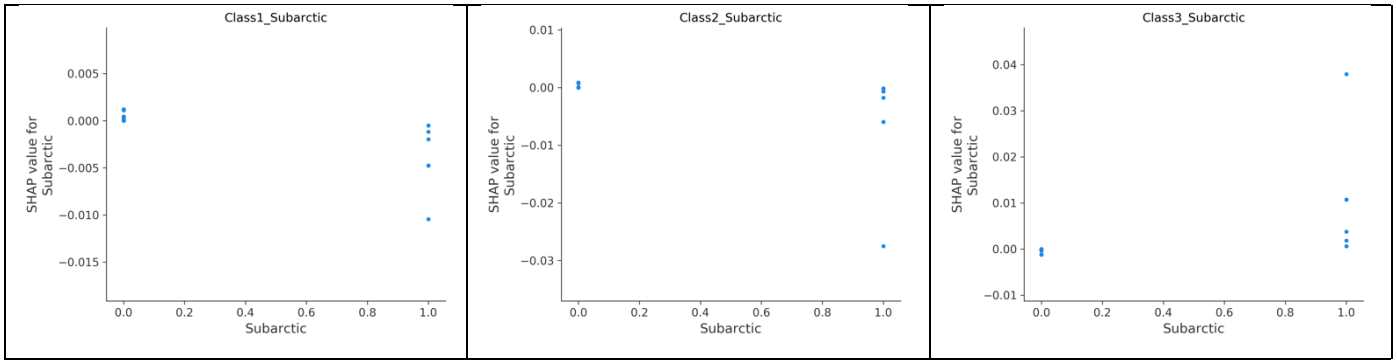

Figure A1 Plots showing SHAP values for different values of the parameters among three classes

**Author Contributions:**

Conceptualization, Ziyadidegan. S., Razavi. M., and Pesarakli. H.; methodology, Ziyadidegan. S., Razavi. M., and Pesarakli. H., Javid. A.; software, Ziyadidegan. S., Razavi. M., and Pesarakli. H.; formal analysis, Ziyadidegan. S., Razavi. M., and Pesarakli. H., Javid. A.; data curation, Ziyadidegan. S., Razavi. M., and Pesarakli. H.; writing—original draft preparation, Ziyadidegan. S.; writing—review and editing, Razavi. M., and Pesarakli. H., Javid. A, Erraguntla. M. "All authors have read and agreed to the published version of the manuscript."

**Data Availability Statement:**

Publicly available datasets were analyzed in this study. This data can be found here:

Data combined and used in this paper by authors:

https://github.com/SamiraZiyadg/COVID-19

Covid Data

https://github.com/nytimes/covid-19-data

Health Data

https://opendata.dc.gov/datasets/1044bb19da8d4dbfb6a96eb1b4ebf629_0?selectedAttribute=ADULT_ICU_BEDS

Demographic, Meteorological, and Location-based data

https://www.kaggle.com/johnjdavisiv/us-counties-covid19-weather-sociohealth-data

**Acknowledgments:**

The first three authors would thank Dr. Darren Homrighausen from Texas A&M University for his valuable feedback on the preliminary draft of this study when he was the instructor of STAT 656: Applied Analytics in 2020.

**Conflicts of Interest:** The authors declare no conflict of interest.